%% file: GraphRiverCast.tex
\documentclass[11pt,a4paper]{article}

\usepackage[top=2.5cm,bottom=2.5cm,left=2.5cm,right=2.5cm]{geometry}

\usepackage[T1]{fontenc}
\usepackage{mathptmx}          
\usepackage{helvet}            
\usepackage{courier}           

\usepackage{amsmath,amssymb,amsfonts}
\usepackage{bm}

\usepackage{graphicx}
\usepackage{booktabs}
\usepackage{multirow}
\usepackage{float}
\usepackage{array}
\usepackage[font=small,labelfont=bf]{caption}
\graphicspath{{figures/}}

\usepackage{authblk}

\usepackage{setspace}
\usepackage[super,sort&compress,comma]{natbib}

\usepackage[colorlinks,linkcolor=blue!70!black,citecolor=blue!70!black,urlcolor=blue!70!black]{hyperref}
\usepackage{url}
\usepackage{titlesec}
\titleformat{\section}{\large\bfseries}{\thesection}{1em}{}
\titleformat{\subsection}{\normalsize\bfseries}{\thesubsection}{1em}{}
\titleformat{\subsubsection}{\normalsize\itshape}{\thesubsubsection}{1em}{}

\usepackage{enumitem}
\setlist{nosep,leftmargin=1.5em}
\usepackage{microtype}
\usepackage{xcolor}

\providecommand{\bibcommenthead}{}

\makeatletter
\newcommand{\bibdataonly}[1]{\if@filesw\immediate\write\@auxout{\string\bibdata{#1}}\fi}
\makeatother

\title{\Large\bfseries Learning global river system hydrodynamics with a topology-informed foundation model}

\author[1,2]{Hancheng Ren}
\author[1,3,$\dagger$]{Gang Zhao}
\author[4]{Shuo Wang}
\author[2]{Louise Slater}
\author[5]{Dai Yamazaki}
\author[6]{Shu Liu}
\author[4]{Jingfang Fan}
\author[2]{Xueying Li}
\author[7]{Shibo Cui}
\author[8]{Ziming Yu}
\author[9]{Shengyu Kang}
\author[1]{Depeng Zuo}
\author[1]{Dingzhi Peng}
\author[1]{Zongxue Xu}
\author[1,$\dagger$]{Bo Pang}

\affil[1]{College of Water Sciences, Beijing Normal University, Beijing, China}
\affil[2]{School of Geography and the Environment, University of Oxford, Oxford, UK}
\affil[3]{Department of Transdisciplinary Science and Engineering, Institute of Science Tokyo, Tokyo, Japan}
\affil[4]{School of Systems Science, Beijing Normal University, Beijing, China}
\affil[5]{Institute of Industrial Science, University of Tokyo, Tokyo, Japan}
\affil[6]{China Institute of Water Resources and Hydropower Research, Beijing, China}
\affil[7]{State Key Laboratory of Hydro-Science and Engineering, Department of Hydraulic Engineering, Tsinghua University, Beijing, China}
\affil[8]{School of Artificial Intelligence, Beijing Normal University, Beijing, China}
\affil[9]{School of Water Resources and Hydropower Engineering, Wuhan University, Wuhan, China}
\affil[$\dagger$]{These authors jointly supervised this work.}

\date{}

\begin{document}

\setstretch{1.3}
\maketitle
\thispagestyle{empty}

\vspace{-2em}
\begin{center}\textbf{Correspondence:} Bo Pang (\href{mailto:pb@bnu.edu.cn}{pb@bnu.edu.cn}) and Gang Zhao (\href{mailto:zhao.g.eb91@m.isct.ac.jp}{zhao.g.eb91@m.isct.ac.jp})\end{center}

\vspace{0.6em}

\begin{center}\textbf{\large Abstract}\end{center}
\vspace{0.4em}
    \noindent
    Accurate river prediction is essential for water, food, and energy security, yet extending it reliably from sparse in-situ gauges to the full river network remains an open challenge. Although machine learning has transformed Earth-system modelling, the same sparse and uneven observations have largely confined its advances for rivers to discrete basins rather than connected river systems. Here we introduce GraphRiverCast (GRC), a topology-informed river foundation model for reach-level prediction across the global river network. Exploiting the connectivity and energetics of rivers, it is trained on sparse gauges and predicts daily multivariate hydrodynamics with no prior river conditions. GRC outperforms strong physics-based and machine-learning models. Attribution experiments show that, compared with standard autoregressive models that rely more on temporal recurrence, GRC reconstructs river states from topology and forcing, learning the dynamics themselves and inheriting the self-correcting character of river systems. Its skill therefore barely decays with lead time, even far beyond its training horizon. It also generalises to ungauged reaches and finer-resolution networks, bringing reliable hydrodynamic prediction to poorly monitored regions. More broadly, harnessing the structure and energetics of a system offers a strategy for machine learning in other observation-limited Earth systems.
\par

\clearpage
\doublespacing


Rivers are the terrestrial conduit of the global water cycle, linking atmosphere, land and ocean\cite{ref44} while sustaining the water, food and energy security of billions\cite{ref42,ref43,ref45,ref46}. Reliable prediction of their hydrodynamics underpins climate adaptation and water management\cite{ref47,ref48}, supports industrial and agricultural production, and enables the mitigation of floods and droughts\cite{ref15,ref49}. Yet predicting the hydrodynamics of the global river system accurately and efficiently at the reach level has remained a defining open challenge of Earth-system science\cite{ref50,ref53}.

Two methodological paradigms have long been pursued. Physics-based models encode the continuity and momentum equations of open-channel flow and deliver mechanistically consistent simulations of the global river system\cite{ref6,ref7,refGloFAS}, but the simplifying assumptions embedded in their governing equations leave persistent systematic errors\cite{ref31,ref53}, and their computational cost limits large-scale calibration and data assimilation\cite{ref50,ref8}. Learning-based methods draw directly on data to predict complex dynamical systems accurately and efficiently, transforming prediction of Earth systems such as the atmosphere and oceans\cite{ref9,ref10,refWenHai}. In hydrology, they achieve remarkable accuracy in well-gauged river basins\cite{refKratzert2018,refReichstein2019,refFengShen2022}, but global rivers still await a comparable advance at the system level. Currently, at the global scale, learning-based river prediction treats each basin as an independent unit\cite{refKratzert2024} and extends to ungauged basins through statistical similarity\cite{refKratzert2019,ref15}, ignoring the connectivity that links neighbouring basins into river systems. At the finer scale of individual reaches, it instead serves as a statistical correction or post-processing of existing river products\cite{refMagni2023,refRiverMamba}, rather than as a self-contained dynamical operator.

The fundamental obstacle to a system-level neural river model is the scarcity of reliable data. Most river reaches have never been gauged, and the gauges that exist are biased towards large rivers and developed regions\cite{ref13,refKrabbenhoft2022}. The narrow, linear geometry of channels makes their hydrodynamic state difficult to retrieve from satellites\cite{ref12,ref38,refDurand2023}. This scarcity simultaneously constrains the development of such a model at both ends, in training and at inference. During training, reaches that have never been gauged provide no direct supervision. At inference, the autoregressive paradigm that is widely used in dynamical-system modelling requires a historical state as its initial field\cite{ref9,ref10}, which most river reaches have never possessed.

A way forward lies in two of the river system's own properties---connectivity and dissipation. Rivers form a natural network structure, a directed, branching topology through which runoff is routed and accumulated among connected reaches\cite{refRI1992}. The state of each reach therefore reflects not only its local, basin-level attributes but also the influence its topology imposes through upstream and downstream connections within the network. This connectivity, in principle, allows sparse gauges to supervise every reach tied to them through the network. Counterintuitively, whether encoding this topology enhances the predictive skill of a neural model remains contested. Some studies report clear gains\cite{ref20,ref21} while others find none from the native network topology\cite{ref23,ref24}, hinting that whether topology helps is conditional on something as yet unidentified.

The second property is dissipation. Initial conditions are often a key factor in simulating dynamical systems, but how strongly a system depends on them is set by its energetics. The balance between dissipation and instability determines whether errors in a system's initial condition are damped or amplified as the simulation evolves\cite{refEckmannRuelle1985}. For instance, the atmosphere is a well-known chaotic system, dominated by instability rather than dissipation, so its butterfly-effect sensitivity amplifies even the smallest errors, requiring precise, high-resolution initial fields for prediction\cite{ref17,ref55}. Rivers, however, are dominated by dissipation, damping perturbations rather than amplifying them\cite{refLighthill}. Errors in a river's initial state therefore shrink rather than grow over time, and the hydrodynamics come to be governed instead by the forcing entering the network and the topology that routes it. In fact, the river's topology itself is the structural product shaped by its dynamical energetics\cite{refRI1992,refMolnar}. A river system can therefore, in principle, be computed without its initial state, as a boundary-value problem in which the state of every reach follows from the forcing and the topology alone. For system-level neural river models this removes the need for initial fields that are hard to obtain globally and replaces them with the readily available topology of the river network.

Building on these two insights into the structure and energetics of river systems, we hypothesise that they jointly resolve the two-sided obstacle identified above. Topology allows sparse gauges to supervise every connected reach at training time, while dissipation allows a model built on that topology to predict without any historical river state at inference, delivering a complete, coherent and accurate hydrodynamic picture of global river systems.

\clearpage
\subsection*{GraphRiverCast}

Here we present GraphRiverCast (GRC), a topology-informed river foundation model that treats the world's rivers as connected systems rather than discrete basins. By combining global river topology with sparse observational supervision from gauges, GRC predicts the hydrodynamics of the full river network without requiring any historical river state for initialisation.

\begin{figure}[H]
\centering
\includegraphics[width=\textwidth]{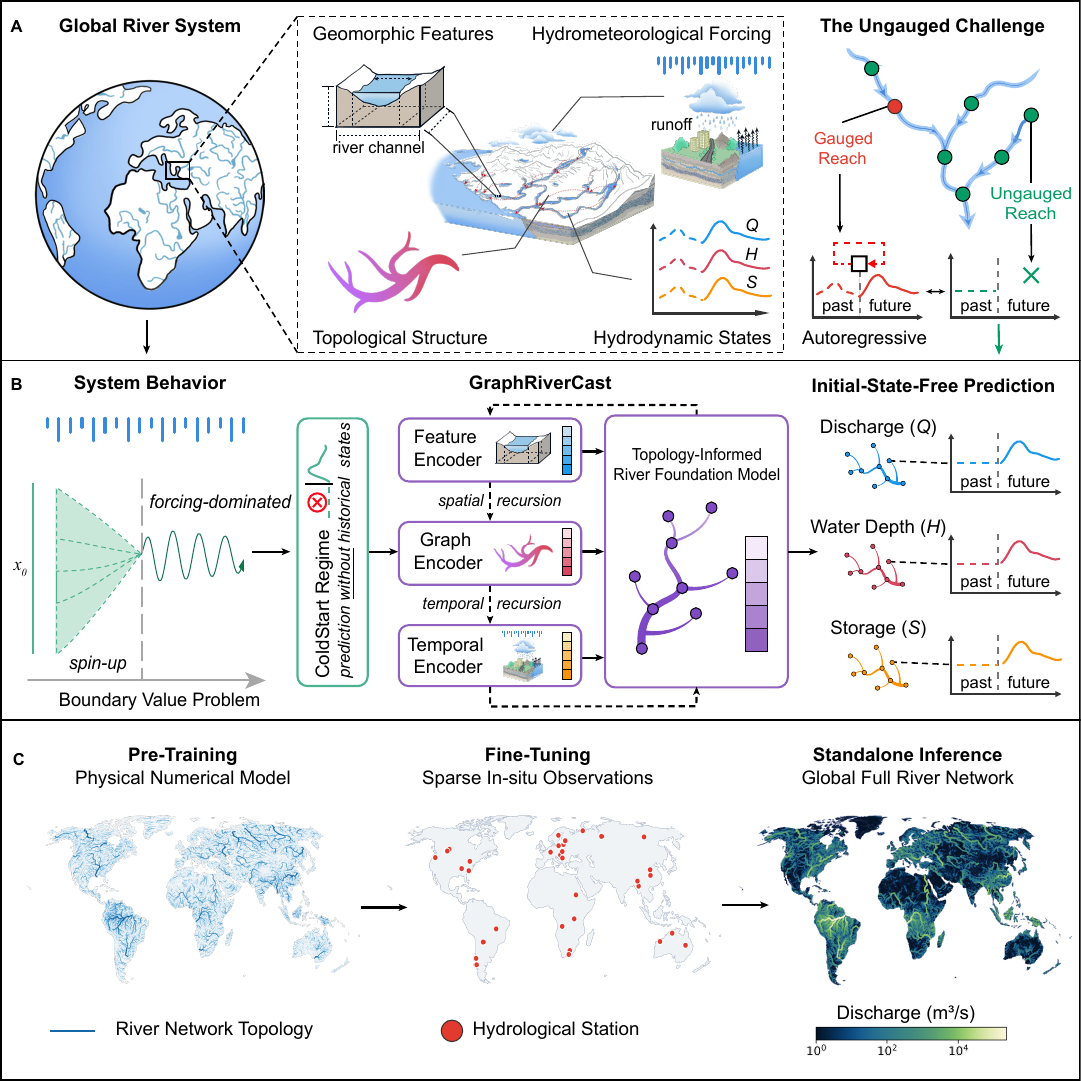}
\caption{\textbf{GraphRiverCast framework and training strategy.}
\textbf{a,} The global river network as a system of connected reaches. Left, the river network represented as a signed, directed graph of hydrologically coherent reaches. Centre, each reach carries geomorphic features, runoff forcing, topology and its hydrodynamic state, namely discharge ($Q$), water depth ($H$) and storage ($S$). Right, at ungauged reaches (green), no historical states exist, so the standard autoregressive paradigm cannot be initialised.
\textbf{b,} Dynamical rationale and architecture. Left, the influence of the initial condition $x_{0}$ decays as trajectories converge towards a common forcing-driven trajectory after spin-up. Reach-scale prediction can thus be treated as a boundary-value problem, which makes the initial-state-free ColdStart regime physically possible. Centre, GraphRiverCast couples a feature encoder, a graph encoder (spatial recursion) and a temporal encoder (temporal recursion) into a topology-informed river model. Right, the fused representation predicts $Q$, $H$ and $S$ at every reach.
\textbf{c,} Training strategy. GraphRiverCast is trained in two stages. It is pre-trained on physical-model simulations over the full network topology, then fine-tuned on sparse in-situ stations (red). The fine-tuned model is then run as a standalone simulator across the global river network, with no observations supplied at inference, shown as a single-time-step discharge snapshot.}
\label{fig:framework}
\end{figure}

GRC is built on a spatio-temporal graph neural network tailored to river systems. It represents the global river system at 15-arc-minute (\textasciitilde25\,km) resolution on a global river map\cite{ref26}, as a signed, directed graph of 127{,}581 hydrologically coherent reaches spanning $\sim$4.4~million~kilometres of channel (Fig.~\ref{fig:framework}a, left). On this graph, GRC predicts the daily hydrodynamic state of every reach, namely discharge ($Q$), water depth ($H$) and channel storage ($S$). Each channel enters the graph as a pair of signed edges, one aligned with the direction of flow and one reversing it, so information can be exchanged both downstream and upstream, with the strength of each edge dynamically modulated by the water-depth gradient between the two reaches.

GRC decomposes the river system into four components (Fig.~\ref{fig:framework}a, centre): the geomorphic features of each channel (width, depth and length, inherited from the physics-based global river model CaMa-Flood\cite{ref6}), the lateral forcing from runoff generated within its contributing catchment\cite{ref28}, the topology connecting each reach to the wider network, and the hydrodynamic state itself. The first three components are available globally from published products. The hydrodynamic state, in contrast, is scarcely observed because most river reaches have never been gauged (Fig.~\ref{fig:framework}a, right). Predicting this state from forcing, geomorphic attributes and topology alone, without supplying a historical state, defines our initial-state-free formulation. For much of the global river network, this is the only feasible prediction regime.

The feasibility of initial-state-free prediction arises from the dissipative dynamics of river systems. Under a common forcing, trajectories from different initial states converge towards the same solution\cite{ref40} (Fig.~\ref{fig:framework}b, left). Long-term reach-level prediction can therefore be approached as a boundary-value problem, governed by external forcing and network structure rather than by an accurately supplied initial state. GRC implements this formulation through three complementary encoders (Fig.~\ref{fig:framework}b, centre). A feature encoder integrates each reach's geomorphic attributes, forcing and the internally propagated hydrodynamic representation carried forward by the model. A graph encoder performs spatial recursion through signed bidirectional message passing over the non-Euclidean river network, while a temporal encoder performs sequential recursion through time. Their outputs are fused into a topology-informed representation that reconstructs the hydrodynamic state of every reach (Fig.~\ref{fig:framework}b, right). Each prediction therefore incorporates information from the connected river network and from the forcing entering its contributing watershed.

Training proceeds in two stages (Fig.~\ref{fig:framework}c). GRC is first pre-trained on global hydrodynamic simulations from CaMa-Flood\cite{ref6}, learning transferable representations of multivariate river dynamics across complete river networks. It is then fine-tuned on sparse in-situ gauges, whose information propagates through the network connectivity to supervise reaches beyond direct observations. The fine-tuned model is then applied across the entire global river network with no gauge data or river state supplied at inference, producing reach-level multivariate predictions even where no gauge exists. Initial-state-free prediction reverses the standard practice of dynamical-system modelling, where many autoregressive models begin from a supplied state field\cite{ref9,ref10}. We therefore evaluate GRC under two regimes, ColdStart and HotStart, which share the same architecture and rollout procedure but differ only in whether a recent river state is provided for initialisation, as described in the following section.

\clearpage
\subsection*{Topology sustains initial-state-free prediction}

We tested the central hypothesis of this work in a controlled experiment, asking whether a river model can predict hydrodynamics ($Q$, $H$ and $S$) without an accurate initial state, and what sustains such prediction if possible. GRC was pre-trained against reference simulations from CaMa-Flood\cite{ref6} to learn an idealised, physically consistent representation of global river hydrodynamics, with each training sample consisting of a historical spin-up followed by a 7-day forward rollout. ColdStart and HotStart use the same GRC architecture and differ only in whether a recent river state is supplied for initialisation. HotStart receives the true most recent state, whereas ColdStart starts from the zero point of the standardised state space. An LSTM widely used in hydrology\cite{ref15,refKratzert2018}, run without an initial state and without access to river topology, serves as a topology-blind counterpart of ColdStart. All models roll forward autoregressively and are evaluated as pseudo-hindcasts over arbitrary lead times under perfect future forcing, isolating each model's intrinsic error accumulation.

\begin{figure}[H]
\centering
\includegraphics[width=\textwidth]{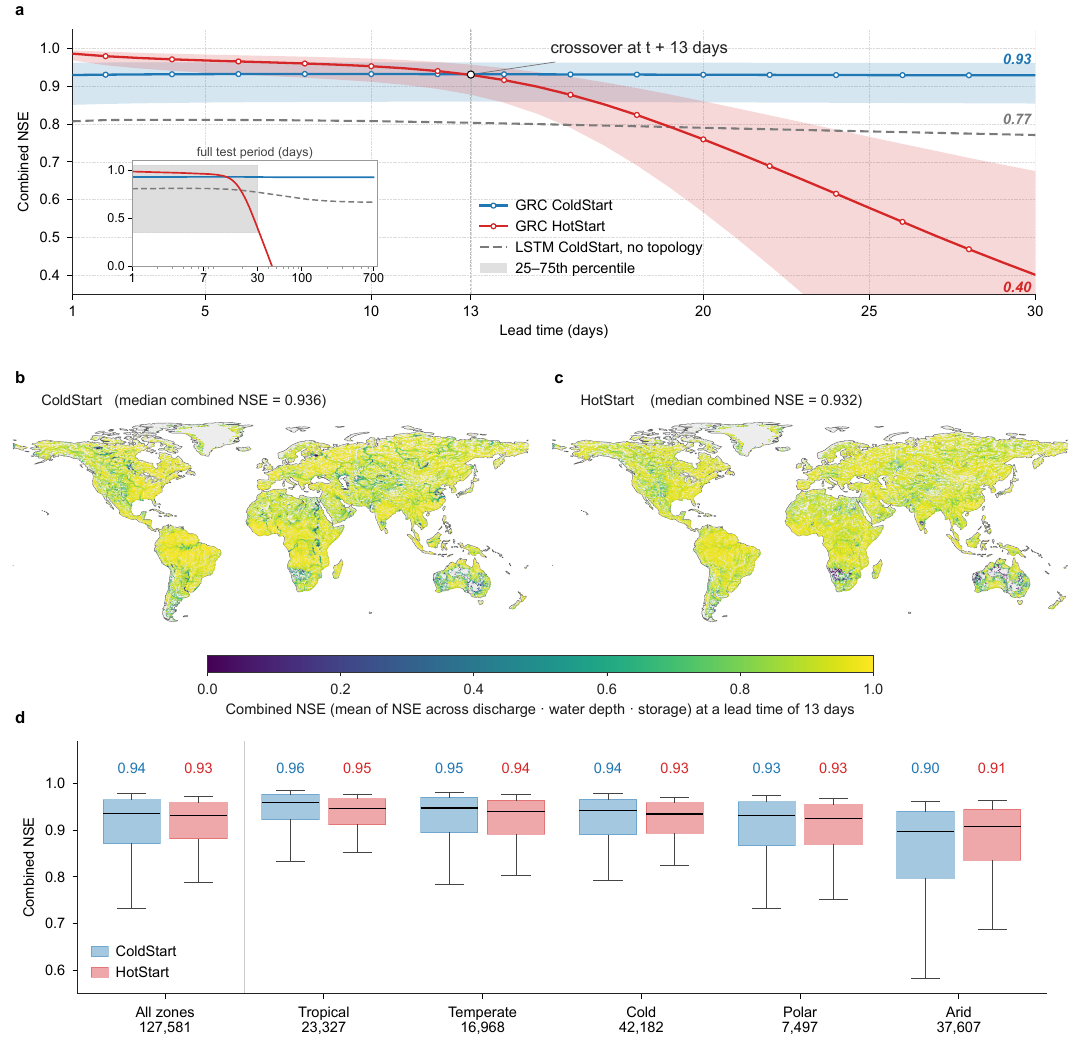}
\caption{\textbf{Lead-time skill of the initial-state-free and state-initialised regimes.}
\textbf{a,} Median combined NSE (the mean Nash-Sutcliffe efficiency across discharge, water depth and storage) against lead time over the held-out test period, for GRC-ColdStart (blue), GRC-HotStart (red) and a topology-free LSTM run initial-state-free (grey dashed), scored against the CaMa-Flood reference. Shaded bands give the 25th--75th percentile across reaches for the two GRC regimes. Inset, the same medians to a 700-day lead on a logarithmic axis, with the 30-day window of the main panel shaded.
\textbf{b, c,} Global maps of combined NSE per reach at a 13-day lead time, for the initial-state-free ColdStart (\textbf{b}) and state-initialised HotStart (\textbf{c}) regimes. Lakes, reservoirs and the Greenland ice sheet are masked, so the map medians differ slightly from the curve medians of panel \textbf{a}, which are over all reaches.
\textbf{d,} Combined NSE by K\"oppen climate zone at a 13-day lead time, with the number of evaluated reaches in each zone below the axis. Boxes give the inter-quartile range and median, whiskers the 10th--90th percentile, and the value above each pair is the median.}
\label{fig:global}

\end{figure}

ColdStart achieves strong predictive skill without any recent river state information. More strikingly, its skill remains almost unchanged across the two-year test period, predicting the 700th day as accurately as the first despite being trained only on 7-day rollouts (Fig.~\ref{fig:global}a, inset). This behaviour contrasts with the error accumulation commonly observed in autoregressive rollouts, where models repeatedly feed their own predictions back as inputs. HotStart, using the same architecture but initialised with the true river state, instead shows the expected degradation over time. Although HotStart is marginally superior at the shortest leads because it benefits from accurate initial conditions, ColdStart overtakes it at day 13 and remains superior thereafter (Fig.~\ref{fig:global}a). The topology-blind LSTM likewise loses skill over long rollouts, whereas ColdStart maintains stable prediction. This contrast indicates that ColdStart's stability is not a generic consequence of starting from an empty state, but emerges from the interaction between initial-state-free prediction and river topology.

The spatial pattern at the day-13 crossover is shown in Fig.~\ref{fig:global}b,c. Across tropical, temperate and cold regions, where runoff forcing is continuous and well resolved, ColdStart consistently outperforms HotStart (Fig.~\ref{fig:global}d). In arid and polar regions, where forcing is sparse or uncertain, this advantage narrows or reverses because the supplied state partially compensates for missing information. ColdStart therefore performs best where forcing provides the strongest dynamical signal.

\begin{figure}[H]
\centering
\includegraphics[width=\textwidth]{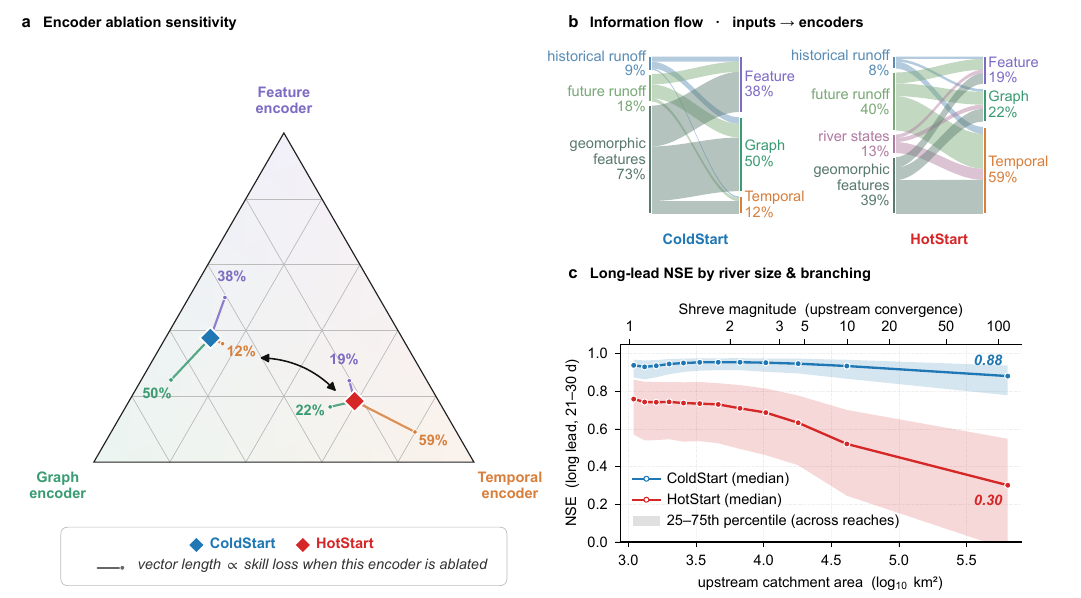}
\caption{\textbf{Encoder dependence and attributed information flow in the two regimes.}
\textbf{a,} Ternary diagram quantifying how much each regime's skill depends on its three encoders, for ColdStart (blue) and HotStart (red). For each encoder we measure an ablation sensitivity, the relative drop in combined NSE when that encoder is removed at inference without retraining. The three sensitivities are normalised to percentages that sum to 100\% and used as the marker's barycentric coordinates, so its position gives the share of skill carried by each encoder. Each share equals the length of the coloured stem reaching towards that encoder's corner and is printed as a percentage at the stem tip. The arrow links the two regimes. Because the ablation is applied at inference without retraining, it upper-bounds an encoder's functional contribution.
\textbf{b,} Sankey diagrams of attributed information flow from input variables (left) to encoders (right) for ColdStart and HotStart, with attribution computed by integrated gradients. Flow width is proportional to the attributed contribution and percentages give each node's share of the total. Inputs are historical runoff, future runoff and geomorphic features, with an additional river-state input for HotStart only.
\textbf{c,} Long-lead (21--30\,d) NSE versus river size and branching, with reaches stratified by upstream catchment area ($\log_{10}\,$km$^{2}$, lower axis) and Shreve magnitude (upper axis), for ColdStart (blue) and HotStart (red). Lines show the median and shaded bands the interquartile range (25th--75th percentile) across reaches in each bin.}
\label{fig:ablation}
\end{figure}

GRC's two regimes differ only in their initialisation, yet their skill curves trace opposite temporal patterns. Attribution experiments reveal that the two regimes learn fundamentally different information strategies (Fig.~\ref{fig:ablation}a). ColdStart relies primarily on the graph representation, whereas HotStart shifts dependence towards temporal recurrence after receiving an explicit river state. The same contrast appears in the input attribution (Fig.~\ref{fig:ablation}b). ColdStart extracts information mainly from geomorphic structure and runoff, while HotStart additionally exploits the supplied river state. This distinction becomes increasingly important with river size and topological complexity, measured as Shreve magnitude\cite{refShreve1966}. At the largest rivers, where long upstream routing pathways converge, HotStart's long-lead skill collapses as its recurrent state drifts, whereas ColdStart remains stable by reconstructing river dynamics from forcing and network structure (Fig.~\ref{fig:ablation}c). These results show that initialisation does not merely improve prediction accuracy but determines whether a model learns transient state corrections or the underlying dynamics that generate those states.

These results reconcile conflicting reports on the value of topology in neural hydrological models, with some studies showing clear benefits from network encoding\cite{ref20,ref21} and others finding limited gains from native topology\cite{ref23,ref24}. A river state is the cumulative outcome of past runoff routed through the network and therefore carries the imprint of the network processes that produced it. When such states are available, explicit topology provides limited additional information because these structural effects are already embedded in the state. When they are unavailable, however, topology becomes the structural basis for recovering how forcing is transformed into river states through the network, learning from topology how strongly each reach is controlled from upstream and from downstream. Retrained ablation experiments further confirm this mechanism, with the graph encoder remaining the dominant source of ColdStart skill and contributing substantially more than in HotStart.

Initial-state-free prediction is therefore not merely feasible but the condition that forces a model to learn river dynamics. By removing the shortcut provided by historical states, ColdStart learns a representation of the dissipative dynamics of the river system rather than a dependence on transient conditions. This learned representation explains its stable long-horizon behaviour and enables simulation across reaches that lack historical states. What data scarcity first imposed as a limitation therefore emerges as a design principle for neural river models.

\clearpage
\subsection*{Fine-tuning against a real-world benchmark}

GRC is pre-trained on abundant physics-based simulations, which provide dynamical consistency but also inherit the systematic biases of the simulation source. Fine-tuning on sparse real observations allows the model to correct these biases and align its predictions with observed river behaviour. The pre-trained GRC was fine-tuned on daily in-situ discharge observations from the Global Runoff Data Centre (GRDC), operated in the initial-state-free ColdStart regime as a standalone simulator, and evaluated on a held-out test period using NSE, Kling-Gupta efficiency (KGE), normalised RMSE (NRMSE) and percent bias (PBIAS). We compared GRC against three leading global models: the grid-based GloFAS-ERA5 reanalysis\cite{refGloFAS}, the vector-based CaMa-Flood hydrodynamic model\cite{ref6}, and the widely validated LSTM benchmark\cite{ref15,refKratzert2018,refKratzert2019}.

\begin{figure}[H]
\centering
\includegraphics[width=\textwidth]{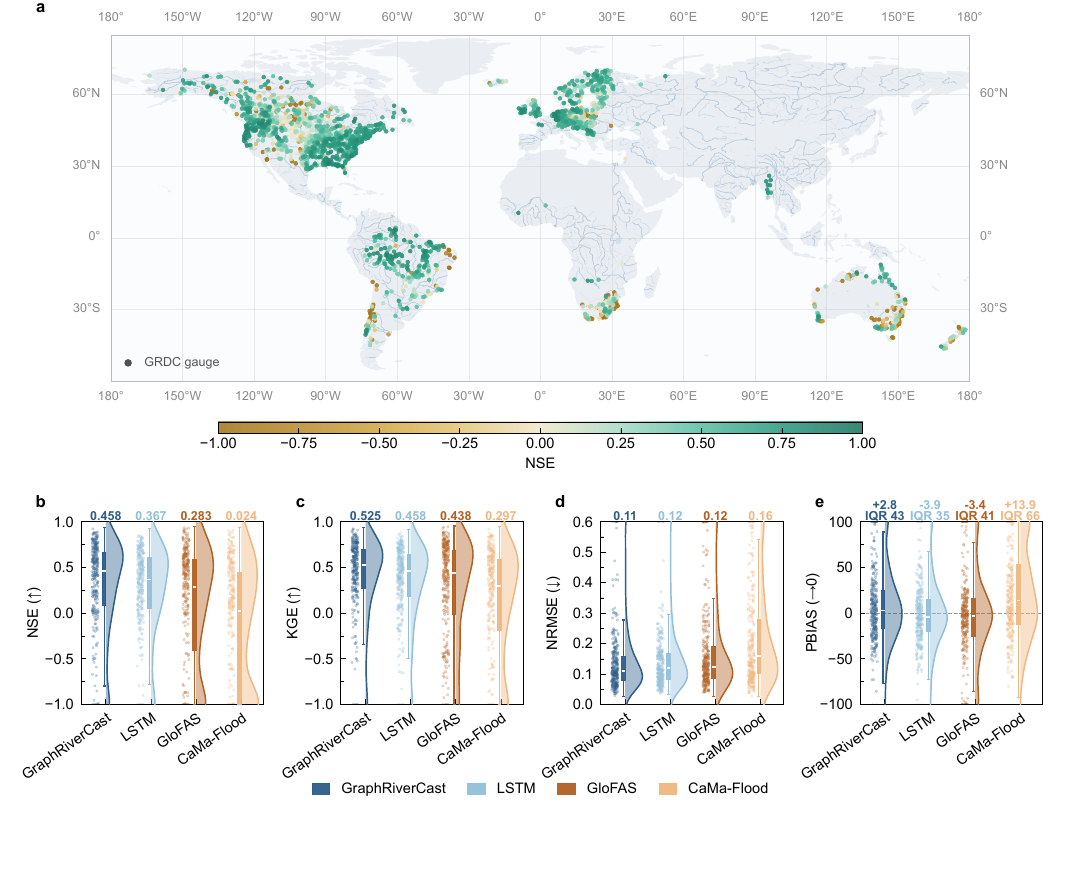}
\caption{\textbf{GraphRiverCast benchmarked against leading physics-based and learning-based global hydrodynamic models.}
\textbf{a}, Global distribution of per-gauge Nash-Sutcliffe efficiency (NSE) for GraphRiverCast over the held-out evaluation period (2018--2019), at the $n=1{,}996$ evaluated in-situ gauges. Marker colour denotes NSE (teal for higher skill, brown for negative), over the global river network (light blue).
\textbf{b}--\textbf{e}, Per-gauge comparison of four metrics across the same gauges for GraphRiverCast, an LSTM benchmark, and the physics-based GloFAS and CaMa-Flood, shown as raincloud plots, each combining jittered points (one per gauge), a box for the interquartile range (IQR) with a median line, whiskers at $1.5\times$IQR and a half-violin kernel density. The value above each distribution is its median, and in \textbf{e} both the median and IQR of PBIAS are given. Higher values indicate better performance for NSE and KGE, lower values for NRMSE, and values near zero for PBIAS. An arrow in each panel marks this direction, upward in \textbf{b} for NSE and \textbf{c} for the Kling-Gupta efficiency (KGE), downward in \textbf{d} for the min-max normalised RMSE, and towards zero in \textbf{e} for the percent bias (PBIAS).}
\label{fig:benchmark}
\end{figure}

Fine-tuned on the same sparse in-situ observations, GRC consistently outperforms all three benchmarks in predictive skill (Fig.~\ref{fig:benchmark}b--e). Its median NSE reaches 0.458, exceeding the LSTM benchmark (0.367) by 25\%. The same advantage is observed for KGE (0.525) and NRMSE (0.11), while PBIAS remains close to zero (+2.8\%), indicating that fine-tuning effectively reduces biases inherited from the simulation pre-training. The improvement is not limited to average conditions. GRC also predicts annual maximum discharge and high-flow volume more accurately than the LSTM benchmark, and its skill remains stable across lead times on real observations. Spatially, the strongest performance occurs in densely monitored temperate basins, whereas larger uncertainties remain in arid and strongly seasonal regions with limited observations (Fig.~\ref{fig:benchmark}a).

The advantage of GRC over data-driven baselines arises from how it organises sparse observations. Basin-based models can share parameters across basins\cite{ref23,refKratzert2024}, but each basin remains hydrologically isolated and cannot directly use observations from connected reaches. GRC instead represents river networks as connected dynamical systems, allowing information from each observation to influence the reaches linked through the network. A reach is therefore predicted not only from its local forcing and attributes, but also from the upstream and downstream structure through which runoff is routed. Connectivity allows sparse gauges to reinforce one another rather than merely accumulate, transforming isolated records into a mutually informing network of constraints. This network-level advantage is strongest where routing interactions are most complex, with GRC showing the largest gains over the LSTM at confluences, where multiple upstream signals must be integrated.

\clearpage
\subsection*{Spatial transfer and scale adaptation of a river foundation model}

Most river reaches worldwide are ungauged, and many applications require predictions at resolutions different from those used for global modelling. A river foundation model should ideally transfer beyond the locations and scales encountered during training. GRC is designed for this purpose. A single global pre-training learns a transferable representation of river dynamics that can be adapted to new regions and resolutions without rebuilding the model. We first test spatial transfer to ungauged reaches and then evaluate scale adaptation on a finer river network unseen during training.

\begin{figure}[H]
\centering
\includegraphics[width=\textwidth]{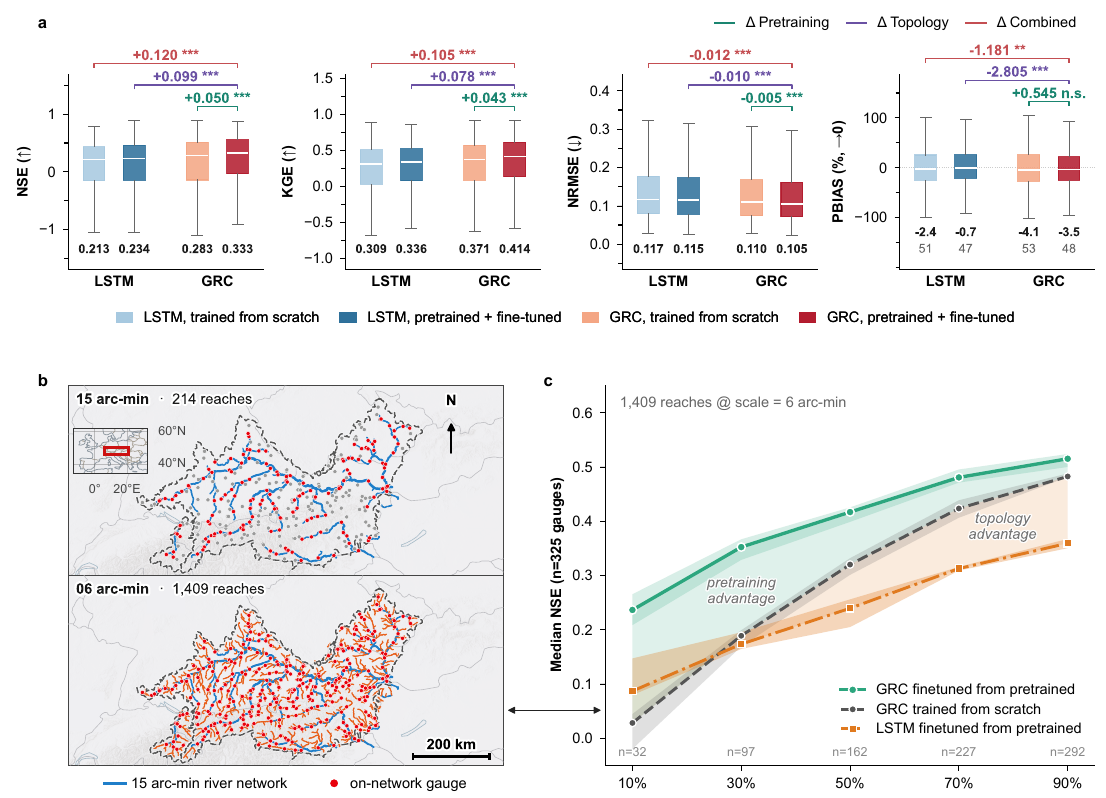}
\caption{\textbf{Spatial and scale generalisation of GraphRiverCast.}
\textbf{a,} Per-gauge skill on ungauged reaches, from five-fold cross-validation over the $1{,}996$ evaluable global GRDC gauges, with each gauge held out once. The four boxes compare the LSTM and GRC, each either trained from scratch or pre-trained then fine-tuned (colours as in the legend), across the four metrics (NSE, KGE, NRMSE, PBIAS). Each box is labelled with its median. Brackets give the difference in medians between GRC (pre-trained and fine-tuned) and three references, namely the pre-trained LSTM ($\Delta$\,Topology), GRC from scratch ($\Delta$\,Pre-training) and the LSTM from scratch ($\Delta$\,Combined). Asterisks denote the statistical significance of the paired per-gauge differences (Wilcoxon signed-rank test, with n.s. denoting not significant).
\textbf{b,} The LamaH river network over the Danube at two resolutions, 15~arc-min (214 reaches, top) and 6~arc-min ($1{,}409$ reaches, bottom). Reaches are blue where resolved at 15~arc-min and orange where resolved only at 6~arc-min. On the coarse map, on-network gauges (red) map onto a modelled reach, whereas off-network gauges (grey) have no reach at this resolution and are recovered only at 6~arc-min. Of the $1{,}409$ reaches at 6~arc-min, the 325 that carry a LamaH gauge are used for evaluation in \textbf{c}.
\textbf{c,} Median NSE over these $n=325$ Danube gauges versus the percentage of gauges used for fine-tuning (the number of supervised gauges is printed beneath each tick), for GRC fine-tuned from the global pre-trained weights (green), GRC trained from scratch (grey, dashed) and an LSTM fine-tuned from pre-trained weights (orange). The two shaded elements differ. The band around each curve gives the min--max across three random gauge-subsampling seeds. The two filled regions between the curves are labelled in the panel.}
\label{fig:generalization}
\end{figure}

Both topology and global pre-training independently improve transfer to ungauged reaches, and their combination produces the largest gain (Fig.~\ref{fig:generalization}a). We assess this generalisation by splitting the global GRDC gauges into five folds and holding each fold out in turn, so every gauge serves once as an unseen reach during evaluation. The LSTM benchmark is evaluated under the same protocol, and both architectures are trained either from scratch or from pre-trained weights. GRC achieves a median ungauged NSE of 0.333, improving by 0.120 over the LSTM trained from scratch (0.213). When pre-training is held fixed, explicitly encoding topology contributes an additional 0.099 gain over the topology-blind LSTM. When topology is held fixed, global pre-training contributes 0.050 over training from scratch. Both contributions are statistically significant, and the pre-training advantage persists across all five folds. These two sources of transfer are complementary. Topology provides the structural pathway for information exchange between reaches, while pre-training learns how river dynamics propagate through this structure.

We next test whether the learned representation transfers across spatial scales. GRC pre-trained on the 15 arc-min global river network is transferred to a finer 6 arc-min network over the Danube (Fig.~\ref{fig:generalization}b), where the LamaH dataset\cite{refLamaH} provides dense, high-quality observations. Many of these gauges are unresolved in the coarser global network and therefore represent a genuine scale-transfer challenge rather than interpolation within the training reaches. Fine-tuning is performed with nested subsets of 10--90\% of local gauges, and performance is evaluated over $n=325$ gauged reaches.

The globally pre-trained GRC consistently leads across all levels of data availability and achieves the largest advantage when local observations are most limited (Fig.~\ref{fig:generalization}c). With only 10\% of gauges available ($n=32$), GRC reaches a median NSE of approximately 0.24, compared with approximately 0.03 for GRC trained from scratch and approximately 0.09 for the pre-trained LSTM. As local observations increase, the relative contributions of pre-training and topology change. When data are scarce, pre-training provides the dominant advantage by transferring global routing experience to regions with limited local information. As observations become denser, topology becomes increasingly valuable because each gauge constrains neighbouring reaches through the connected network.

This scale transfer arises because GRC learns river routing on the topology of the network rather than on a particular spatial discretisation. River networks exhibit hierarchical and approximately scale-invariant structural organisation\cite{ref37}, so the relationships governing runoff propagation persist across resolutions. The pre-trained model therefore transfers a learned routing process rather than resolution-specific patterns, retaining its advantage even on finer networks absent from the original global representation. A single globally trained GRC can consequently be adapted to new resolutions and local regions, transferring the routing knowledge of the global river system to where predictions are most needed.

\clearpage
\section*{Discussion}

There is growing expectation that river prediction should achieve the system-level capability of modern weather forecasting, providing a complete and coherent representation of global hydrodynamics. We have introduced GraphRiverCast (GRC), a river foundation model that treats the world's rivers as connected dynamical systems and predicts reach-level, multivariate hydrodynamics across the full river network. By learning a self-contained hydrodynamic operator rather than a task-specific correction, GRC extends learning-based river prediction beyond individual basins towards a system-level representation of global river dynamics. More broadly, GRC provides a test case for how machine learning can learn the underlying dynamical processes of sparsely observed physical systems by harnessing their energetic properties and physical structure.

Providing previous states of the variable being predicted can improve short-term forecasting, but it can also discourage a model from learning the dynamics that generate those states. Previous target states provide a compressed record of past system evolution, allowing a model to exploit transient information rather than infer the processes that produced it. Many successful forecasting systems provide such state information as inputs, following the paradigm of chaotic systems such as the atmosphere, where accurate initial conditions are essential because perturbations amplify over time. Rivers represent a different dynamical regime. Their strong dissipation progressively removes memory of initial states, leaving long-term behaviour governed by external forcing and the structure through which that forcing propagates.

Our results show that this distinction shapes what a neural river model learns. When previous river states are supplied, GRC relies more strongly on temporal recurrence and learns a state-conditioned correction strategy. Removing this target-state shortcut shifts the model towards the forcing and network structure that generate river states, allowing it to recover the intrinsic dynamics of the system. The resulting representation captures the dissipative and self-correcting behaviour of rivers and avoids the progressive degradation associated with autoregressive error accumulation. This suggests that the design of learned dynamical models should consider not only the architecture used, but also the information pathway through which the model accesses the system.

This perspective also helps explain conflicting reports on the value of topology in neural hydrological models\cite{ref20,ref21,ref23,ref24}. A river state is the cumulative outcome of past runoff routed through the network and therefore carries the imprint of the structural processes that produced it. When previous target states are available, explicit topology can appear redundant because much of the information generated by connectivity is already embedded in the state. When those states are unavailable, however, topology becomes the physical basis for recovering how forcing is transformed into river states across the network. The contribution of physical structure therefore depends on the information already available to the model. This phenomenon is not unique to river systems. Similar effects may explain why graph representations have produced inconsistent gains in broader graph-based temporal learning tasks\cite{refEdgeBank2022,refTGB2023,refRecurrency2024,refSTID2022}, where the target's own history or node identity may already encode much of what the graph would supply. Graph structure may reveal its contribution most clearly when a model is compelled to learn from the underlying physical organisation of the system rather than from supplied states.

The significance of learning intrinsic river dynamics extends beyond predictive accuracy. Because GRC needs no historical river states at inference and is not a correction layered on an existing river product, it can provide hydrodynamic predictions in regions where conventional approaches remain constrained by limited observations and monitoring capacity. This capability is particularly important for under-monitored regions, where the need for reliable flood and water-resource information is often greatest but observational infrastructure is weakest. By learning a global representation of river dynamics, GRC provides a foundation on which regional predictions can be developed without requiring each basin to independently build dense monitoring networks or task-specific models.

Despite these advances, several challenges remain in extending this framework towards fully operational global river prediction. First, human regulation of rivers is not yet explicitly represented. Dams and reservoirs alter both river states and network connectivity\cite{refGrill2019}, but their operational decisions are often unavailable, inconsistent across regions, or difficult to represent within a unified global framework. Incorporating such interventions as dynamic changes to the river network remains an important challenge for future river foundation models. Second, although GRC suppresses internally generated error growth, its remaining uncertainty is increasingly determined by the external forcing that drives the system. Uncertainties in weather prediction and land-surface modelling therefore become the dominant sources of long-range prediction uncertainty. Improving these Earth-system components and developing uncertainty-aware forecasting frameworks will be essential for extending the useful horizon of global river prediction. Third, topology can propagate information from sparse observations but cannot fully replace observations in completely isolated subnetworks. Where no observational anchor exists, prediction must rely more heavily on learned structural similarity rather than direct network-based information transfer.

Looking forward, coupling GRC with operational weather forecasts, assimilating emerging observations such as SWOT water-surface elevations\cite{ref38} and GRACE/GRACE-FO terrestrial water storage\cite{refTapley2019,ref39}, and representing human interventions as dynamic modifications of the river network could further improve real-time and uncertainty-aware river forecasting. More broadly, the strategy demonstrated here may extend beyond rivers to other observation-limited physical systems, where key dynamical states are difficult to measure but structural information and physical constraints remain available. Harnessing structure and energetics in place of states that cannot be observed may therefore offer a broader strategy for machine learning models of complex dynamics.

\input{refs_main}

\section*{Acknowledgements}
We thank the developers of CaMa-Flood for making their global river hydrodynamic model publicly available, and P.\ Lin for providing the GRADES global runoff dataset. We acknowledge the Global Runoff Data Centre (GRDC) for providing in-situ discharge observations used in model fine-tuning and evaluation. We also thank the developers of the LamaH-CE dataset for providing high-quality catchment data for the Upper Danube Basin experiments. We thank Ana Miguel Silva Tavares for help with the visual design of the figures.

\end{document}